\documentclass[letterpaper,10pt,conference]{ieeeconf}
\IEEEoverridecommandlockouts
\usepackage[utf8]{inputenc}
\usepackage[T1]{fontenc}
\IfFileExists{ieeeconf.cls}{}{\usepackage{lmodern}}
\usepackage{cite}
\usepackage{amsmath,amssymb,amsfonts,bm}
\usepackage{graphicx}
\usepackage{booktabs,multirow,tabularx,array,float}
\usepackage{dblfloatfix}
\usepackage{placeins}
\usepackage[table,xcdraw]{xcolor}
\usepackage{mathtools}

\usepackage[font=footnotesize,labelfont=footnotesize]{caption}
\makeatletter
\let\NAT@parse\undefined
\makeatother

\usepackage[colorlinks=true,citecolor=blue,urlcolor=blue,linkcolor=blue]{hyperref}
\usepackage[nameinlink,noabbrev]{cleveref}

\newcommand{\pcell}[3]{%
  \shortstack{#1\\[-0.35ex]{\tiny[#2,#3]}}%
}

\newcommand{\bestpcell}[3]{%
  \shortstack{\textbf{#1}\\[-0.35ex]{\tiny[#2,#3]}}%
}

\newcommand{\secondpcell}[3]{%
  \shortstack{\underline{#1}\\[-0.35ex]{\tiny[#2,#3]}}%
}

\newcommand{\mcell}[3]{%
  \shortstack{$#1$\\[-0.35ex]{\tiny$[#2,#3]$}}%
}
\newcommand{\method}{\mbox{MoE-DAgger}}

\newcommand{\ind}{\mathbb{I}}

\newcommand{\NumViews}{4}

\newcommand{\NumExperts}{5}

\newcommand{\ChunkLength}{30}
\newcommand{\PolicyRate}{20\,Hz}

\newcommand{\GreenEntropyThreshold}{0.50}
\newcommand{\RedEntropyThreshold}{0.46}
\newcommand{\GreenACTThreshold}{0.78}
\newcommand{\RedACTThreshold}{0.47}
\newcommand{\GreenEnsembleThreshold}{1.11}
\newcommand{\RedEnsembleThreshold}{0.91}

\newcommand{\EntropyWindow}{5}
\newcommand{\EntropyPersistence}{5}
\newcommand{\MinimumEntropyStep}{5}
\newcommand{\OffPathThreshold}{0.15}
\newcommand{\TransitionStartProb}{0.10}
\newcommand{\TransitionConfirmNext}{0.60}
\newcommand{\TransitionConfirmPrev}{0.35}
\newcommand{\TransitionConfirmSteps}{3}
\newcommand{\TransitionCancelProb}{0.05}
\newcommand{\TransitionCancelSteps}{4}
\newcommand{\GreenTransitionTimeout}{30}
\newcommand{\RedTransitionTimeout}{30}
\newcommand{\RegressionPersistence}{3}
\newcommand{\HomeArmRadius}{0.17\,m}
\newcommand{\HomeRadius}{0.15\,m}

\newcommand{\BudgetTolerance}{5\%}

\newcommand{\TotalRollouts}{800}

\newcommand{\HeadlineTaskSuccess}{22/26 percentage points}

\newcommand{\TraceAlignedRateACT}{70.0/72.0\%}
\newcommand{\TraceAlignedRateEnsemble}{63.3/60.0\%}
\newcommand{\TraceAlignedRatePhaseUnaware}{3.3/2.0\%}
\newcommand{\TraceAlignedRateOurs}{83.3/80.0\%}

\title{\LARGE\bf
MoE-DAgger: Transition-Aware Router-Entropy Gating \\ for Selective Human Correction in Phase-Structured Manipulation}

\author{Sungjun Heo$^{1}$, Sungjun Yang$^{2}$, Heedon Jeong$^{3}$, Junhyoung Ha$^{2, \dagger}$ and Jeong hwan Jeon$^{1, 3, \dagger}$
\thanks{This work was supported by the InnoCORE program of the Ministry of Science and ICT (\#N10260008 and UNIST 1.260037.01), by the National Research Foundation of Korea (NRF) grant funded by the Korean government (MSIT) (No. RS-2022-NR070854), and by the Institute of Information \& communications Technology Planning \& Evaluation (IITP) grants funded by the Korean government (MSIT) (No. RS-2020-II201336, Artificial Intelligence Graduate School Program (UNIST); No. RS-2026-25527734, Development of an industry-specified multimodal hyperscale foundation model).}
\thanks{$^{\dagger}$ Equal Advising, Corresponding authors: \{j.ha, jhjeon\}@unist.ac.kr}%
\thanks{$^{1}$Artificial Intelligence Graduate School, $^{2}$Department of Mechanical Engineering, and $^{3}$Department of Electrical Engineering, Ulsan National Institute of Science and Technology (UNIST), Republic of Korea.}%
}

\date{}

\ifdefined\pdfsuppressptexinfo
\fi

\ifdefined\pdfinfoomitdate
\fi

\ifdefined\pdftrailerid
  \pdftrailerid{}
\fi

\begin{document}
\maketitle
\thispagestyle{empty}
\pagestyle{empty}

\begin{abstract}
Selective data acquisition in Dataset Aggregation (DAgger) on robots requires deciding which learner rollouts
to correct and when to intervene. In manipulation with ordered phases, router
entropy from a phase-supervised soft mixture-of-experts (MoE) is insufficient:
the same value can reflect either a normal adjacent-phase handoff or support for nonadjacent or earlier phases. We propose \method{}, which uses causal
router history to distinguish these cases without providing the policy with a phase label, progress variable, or hard phase command during execution. The acquisition rule allows an adjacent transition to settle, but requests
correction when high entropy persists outside such a transition, the transition
times out, or routing persistently favors an earlier phase. Under matched active
human-control budgets, \method{} achieves 84.0\% task success on Green and
78.0\% on Red, exceeding phase-unaware filtered entropy by 22 percentage points
on Green and 26 points on Red, with no observed collisions in 50 rollouts per
task. On 50 shared intervention-free traces per task, 83.3\% and 80.0\% of its
first proposals fall on failed rollouts within predefined failure windows when
15 and 25 traces are selected, compared with 70.0\% and 72.0\% for temporal
action-chunk disagreement and 63.3\% and 60.0\% for ensemble disagreement.
Shadow replay and an active no-veto ablation indicate that suppressing queries
at normal adjacent handoffs is a key contributor to the improved allocation. 

\par\centerline{Project page: \href{https://SJ824.github.io/MoE-DAgger/}{https://SJ824.github.io/MoE-DAgger/}}
\end{abstract}

\vspace{-0.1em}

\section{Introduction}
\label{sec:introduction}

Visuomotor behavior cloning can reproduce long-horizon manipulation from
demonstrations~\cite{zhao2023act,chi2023diffusion}, but compounding errors can
move a deployed policy beyond the demonstrated state distribution. DAgger addresses this shift by adding expert
supervision on learner-visited states~\cite{ross2011dagger}; with intermittent human takeover, however, monitoring and corrective control
are costly in real-world interactive
systems~\cite{kelly2019hgdagger,hoque2022thriftydagger,
spencer2020interventions,chisari2022correctme}. Under a fixed budget of active human control, acquisition must choose both
which rollout to correct and when to take over. Rollout-level selection alone
is insufficient: a proposal after collision onset is too late, whereas one at
a benign handoff prevents autonomy from reaching later interaction states. We
therefore evaluate selective acquisition by rollout selectivity,
failure-mode-specific critical-window alignment, and burden on successful
rollouts.
\begin{figure}[t]
    \centering
    \includegraphics[width=0.99\columnwidth,
    trim={14pt 245pt 573pt 7pt},clip]{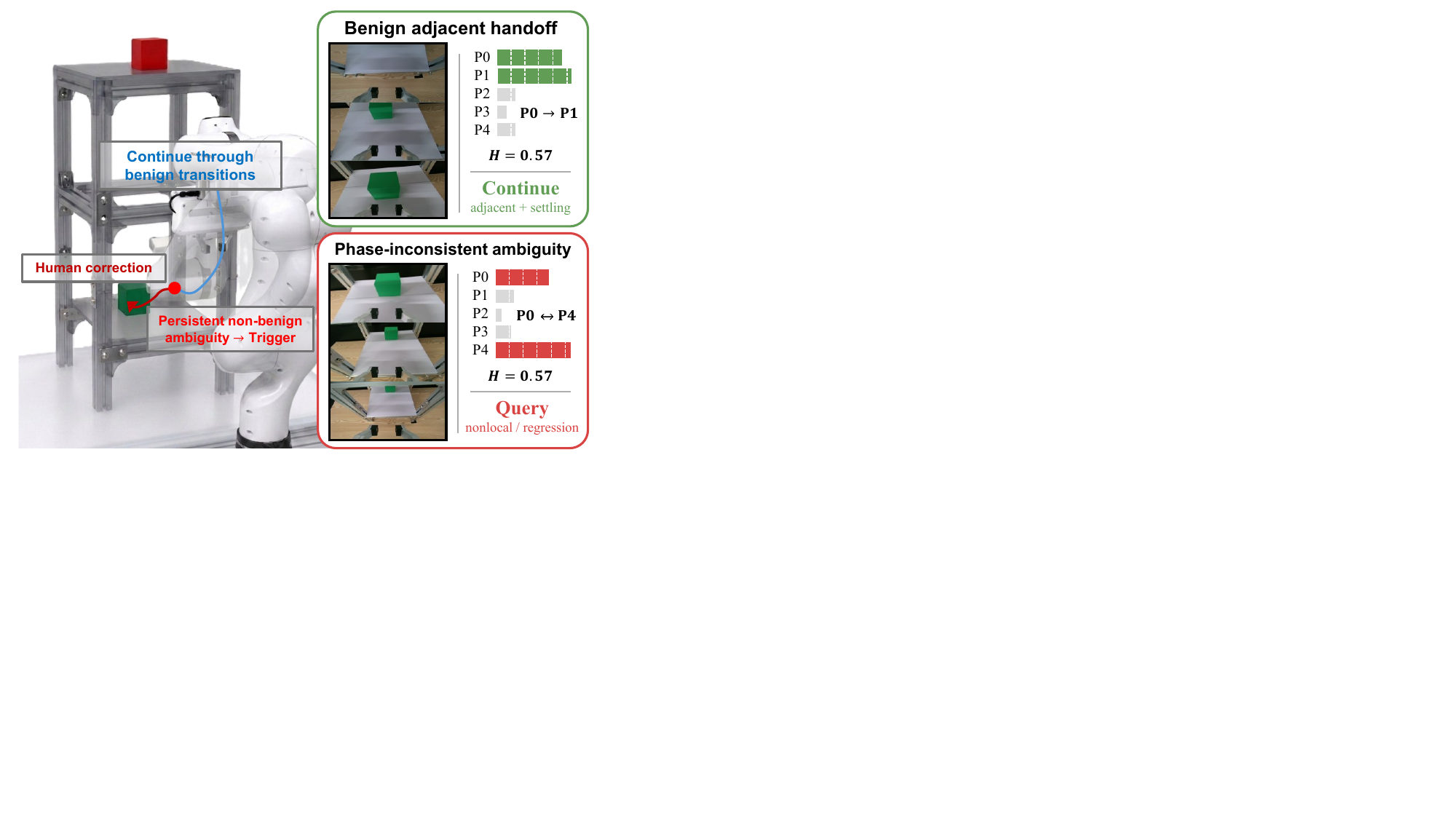}
    \caption{\textbf{Equal router entropy can require different acquisition
decisions.}
\emph{Green:} The robot remains autonomous while router support shifts from
$P0$ to $P1$ and settles within the allowed transition time.
\emph{Red:} The same entropy value can arise when the router supports the
nonadjacent phases $P0$ and $P4$. The acquisition rule does not issue a query
during the settling adjacent transition, but requests human correction if the
nonadjacent support persists.}
    \label{fig:overview}
\end{figure}
Our demonstrations follow a known ordered semantic phase graph and supervise
phase-anchored ACT-style action heads within a soft
MoE~\cite{puigcerver2024softmoe,mazza2026supervisedmoe}. Each head predicts a
complete Action Chunking with Transformers (ACT)~\cite{zhao2023act} chunk rather
than a hand-authored skill or finite-state mode. During execution, routing
remains soft and observation-driven; phase topology is used only by the
acquisition rule. Consequently, the same scalar entropy can describe either a
normal current-to-successor transition or support for nonadjacent or earlier
phases (\Cref{fig:overview}).

\method{} uses only causal router history to track legal current-to-successor
support and temporal settling. It suppresses queries at bounded compatible
handoffs and triggers on persistent non-benign ambiguity, transition timeout,
or regression. The phase graph is therefore an acquisition prior, not a runtime action-routing constraint.
We evaluate (i) downstream utility from independently acquired correction sets
under matched active human-control time and (ii) count-matched first-proposal
allocation on the same 50 Green and 50 Red intervention-free traces. The latter credits a failed rollout only when its first proposal lies in the
corresponding fixed failure-mode-specific critical window. 
Phase-unaware filtered-H is the primary same-signal mechanism ablation, while ACT-Temporal-Disagreement and Ensemble-Disagreement are the two principal phase-unaware action-space baselines. Across the two tasks, \method{} improves
task success over the same-signal gate by \HeadlineTaskSuccess{} and achieves
the highest aligned yield in all four task--budget blocks. Shadow replay and an
active no-veto test isolate the adjacent-handoff mechanism.

\noindent\textbf{Contributions.}
We (1) show that scalar router entropy does not determine transition structure
in an unclamped phase-supervised soft MoE; (2) introduce a causal history-only
gate that conditions persistent entropy on legal support and settling while
retaining timeout and regression triggers; and (3) evaluate downstream utility,
critical-window-aligned allocation, and the adjacent-veto mechanism under matched
human-control time.

\section{Related Work}
\label{sec:related_work}
\subsection{Interactive Imitation and Selective Acquisition} DAgger aggregates expert supervision on learner-induced states~\cite{ross2011dagger, kim2026deployment}. SafeDAgger~\cite{zhang2017safedagger}, HG-DAgger~\cite{kelly2019hgdagger}, EnsembleDAgger~\cite{menda2019ensembledagger}, and ThriftyDAgger~\cite{hoque2022thriftydagger} reduce supervision using learned safety, human takeover, disagreement, novelty, or risk, while related systems learn from interventions or mixed corrective/evaluative feedback~\cite{spencer2020interventions, chisari2022correctme}. Recent work gates on diffusion uncertainty or learned intervention criteria, or changes correction granularity through skill-level queries, residual policies, and force-aware editing~\cite{he2025uncertainty,cai2025aim,luijkx2025askdagger,huang2026terdagger}. AutoIntervene compares proposed chunks with a phase-local visual--action memory and calibrates bidirectional policy--operator transfer~\cite{tang2026autointervene}. Our gate instead uses soft-MoE router history and a known phase graph to
distinguish nominal adjacent handoffs from correction-worthy ambiguity.

\vspace{-0.5em}

\subsection{Action-Sequence Policies and Uncertainty} ACT and diffusion policies model long-horizon, multimodal visuomotor behavior~\cite{zhao2023act,chi2023diffusion}. Deep ensembles estimate predictive disagreement~\cite{lakshminarayanan2017deepensembles}, and Diff-DAgger derives uncertainty from a deployed diffusion policy~\cite{lee2024diffdagger}. We compare phase-unaware temporal disagreement across overlapping
ACT chunks and same-family ensemble disagreement.

\vspace{-0.5em}

\subsection{Phase-Structured Policies and Expert Routing} MoE models learn specialized computation through sparse or soft routing~\cite{shazeer2017moe,puigcerver2024softmoe}. State-gated experts combine supervision sources in contact-rich manipulation~\cite{surendran2026bridge}, while phase-supervised MoEs use semantic phases to specialize ACT-style action prediction~\cite{mazza2026supervisedmoe}. Maneuver-conditioned hierarchical control~\cite{heo2026shiftrl} and stage-conditioned imitation~\cite{ouyang2025scil} provide the policy with a runtime stage or mode command; our acquisition gate uses phase topology only to interpret router ambiguity.

\begin{figure}[t]
    \centering
    \includegraphics[width=\columnwidth,
    trim={7pt 170pt 1120pt 12pt},clip]{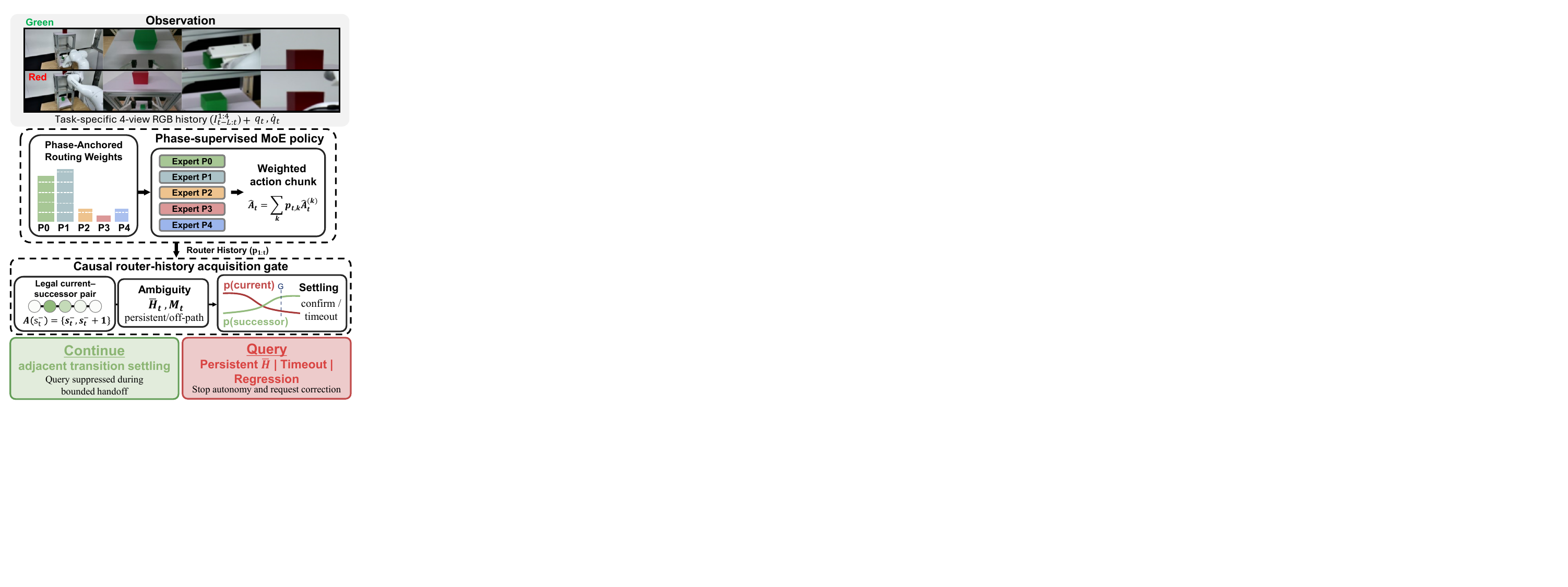}
    \caption{\textbf{Online \method{} acquisition architecture.}
Separate policies for Green and Red map synchronized four-view RGB history and
proprioception to a soft mixture of outputs from five phase-anchored action heads. The
acquisition module receives only causal router history. It permits a bounded,
settling transition between the current phase and its legal successor, but
requests human correction when high entropy persists outside that case, when
the transition times out, or when routing repeatedly moves backward.}
    \label{fig:method_overview}
\end{figure}

\section{Method}
\label{sec:method}

\subsection{Problem Setup and Phase-Supervised Soft MoE}

For task \(d\in\mathcal T\), we train a separate policy \(\pi_\theta^d\) and
annotate demonstrations with the ordered semantic graph
\(0\!\rightarrow\!1\!\rightarrow\!\cdots\!\rightarrow\!S-1\). 
At runtime, the phase graph is used only by the acquisition gate, not to command
the policy: no phase/progress variable or hard mode command is provided to the
policy, and the router remains free to blend or revisit heads. Starting from
\(\mathcal D_0^d\), round \(i\) executes the learner, proposes takeover from
policy-output history, records an accepted suffix \(\mathcal C_i^d\), and
retrains on \(\mathcal D_{i+1}^d=\mathcal D_i^d\uplus\mathcal C_i^d\).
Proposal timing is part of acquisition because an accepted takeover replaces
the remainder of the autonomous rollout with human control, changing which
later learner-induced states are visited and corrected. We therefore evaluate
proposal timing against synchronized physical task semantics rather than
treating the router's head index as ground-truth phase.
At policy step \(t\),
\(\mathbf o_t=(\mathbf I_{t-L:t}^{1:V},\mathbf q_t,\dot{\mathbf q}_t)\)
contains synchronized external/wrist views, task-relevant crops, and
proprioception. In both tasks, the \(S=5\) phase anchors are approach,
entry/alignment, push, withdrawal/clearance, and home return. A shared
encoder produces \(\mathbf h_t\); each of the \(K=S\) phase-anchored heads
predicts a complete ACT chunk
\(\widehat{\mathbf A}_t^{(k)}\in\mathbb R^{C\times n_a}\), and
\begin{equation}
    \scalebox{0.99}{$\displaystyle
    \begin{aligned}
     p_{t,k}
     &=
     \frac{\exp(r_k(\mathbf h_t)/T_r)}
     {\sum_{j=0}^{K-1}\exp(r_j(\mathbf h_t)/T_r)},
     \quad k\in\{0,\ldots,K-1\},\\[-0.2em]
     \widehat{\mathbf A}_t
     &=
     \sum_{k=0}^{K-1}p_{t,k}\widehat{\mathbf A}_t^{(k)}.
    \end{aligned}
    $}
    \label{eq:router}
\end{equation}
We temporally ensemble overlapping chunks for execution.

Each supervised
frame has a manually reviewed phase label \(z_t\); correction suffixes label
only the phases they contain. With label-smoothed target
\(\widetilde{\mathbf y}_t\), training uses

\vspace{-1em}

\begin{equation}
\begin{split}
\mathcal L={}&
\mathcal L_{\rm BC}
+\lambda_r\operatorname{CE}(\widetilde{\mathbf y}_t,\mathbf p_t)
+\lambda_e\sum_k
  \widetilde y_{t,k}\mathcal L_{\rm BC}^{(k)}\\[-0.2em]
&+\lambda_b\mathcal L_{\rm balance}
+\lambda_s\mathcal L_{\rm chunk\text{-}smooth}.
\end{split}
\label{eq:loss}
\end{equation}

Because \(\mathbf p_t\) is not projected onto the phase graph, the router
remains re-entrant and backward routing remains observable to the acquisition
gate.

\subsection{Router Ambiguity and Legal Phase Support}

We compute normalized Shannon entropy and its causal \(W_H\)-step mean,
\begin{equation}
    \scalebox{0.93}{$\displaystyle
     H_t=
     -\frac{1}{\log K}
     \sum_{k=0}^{K-1}p_{t,k}\log p_{t,k},
     \quad
     \bar H_t=
     \frac{1}{W_H}\sum_{\ell=0}^{W_H-1}H_{t-\ell},
     $}
    \label{eq:entropy}
\end{equation}
with clipping only for numerical evaluation of \(\log p_{t,k}\). Entropy is an
ambiguity statistic, not a calibrated failure probability, reward, or safety
certificate. Let \(s_t^-\) be the stabilized phase frozen before the tracker is
updated. Legal adjacent support and off-path mass are
\begin{equation}
  \scalebox{0.86}{$\displaystyle
  \mathcal A(s_t^-) =
  \begin{cases}
    \{s_t^-,s_t^-+1\}, & s_t^-<S-1,\\
    \{S-1\},           & s_t^-=S-1,
  \end{cases}
  \quad
  M_t = 1-\sum_{\mathclap{k\in\mathcal A(s_t^-)}}p_{t,k}.
  $}
  \label{eq:combined_support}
\end{equation}

\noindent\textbf{Proposition 1 (entropy does not determine transition
structure).}
A router supported on a legal pair satisfies
\(H_t\leq\log 2/\log K\). More generally, if the legal pair is top-2 and
\(0\leq M_t=m\leq\tau_M\leq(K-2)/K\), maximizing entropy within the
legal and off-path groups gives
\begin{equation}
    \scalebox{0.98}{$\displaystyle
     H_{\max}^{\rm veto}(m)=
     \frac{h_2(m)+(1-m)\log2+m\log(K-2)}
     {\log K},
     $}
    \label{eq:veto_entropy_bound}
\end{equation}
where
\(h_2(m)=-m\log m-(1-m)\log(1-m)\).
Hence entropy combines legal current--successor sharing, uncertainty about
being on path, and dispersion among off-path heads. For \(K=5\),
\(H_{\max}^{\rm veto}(0)=0.431\) and
\(H_{\max}^{\rm veto}(0.15)=0.731\); thus a locally compatible handoff can
exceed both deployed thresholds once allowed off-path mass is present.
Moreover,
$\mathbf p_{\mathrm{legal}}
=
\left(.475,.475,\tfrac{1}{60},\tfrac{1}{60},\tfrac{1}{60}\right)$
and
$\mathbf p_{\mathrm{nonlocal}}
=
\left(.475,\tfrac{1}{60},\tfrac{1}{60},\tfrac{1}{60},.475\right)$
have identical entropy \(0.567\) for \(s_t^-=0\) but incompatible support.
The phase graph therefore affects only acquisition logic: it determines
whether entropy-based queries can be temporarily suppressed and defines the legal
transitions and phase order used by timeout and regression tests, but never
constrains policy routing.

\subsection{Transition-Aware Query Rule}
\label{sec:gate}

For stabilized phase \(s=s_t^-<S-1\), a current--successor candidate is
locally compatible when
\begin{equation}
 L_t(s)=
 \ind\!\left[
 \operatorname{Top2}(\mathbf p_t)=\{s,s+1\}
 \land M_t\leq\tau_M
 \right].
\label{eq:local_compatibility}
\end{equation}

An inactive candidate opens when \(L_t(s)=1\) and
\(p_{t,s+1}\geq\gamma_{\rm start}\). It confirms when
\(\operatorname*{arg\,max}_k p_{t,k}=s+1\),
\(p_{t,s+1}\geq\gamma_{\rm conf}\), and
\(p_{t,s}\leq\gamma_{\rm prev}\); it cancels when the argmax is \(s\) and
\(p_{t,s+1}\leq\gamma_{\rm cancel}\).
Confirmation and cancellation require \(P_{\rm conf}\) and
\(P_{\rm cancel}\) consecutive frames, respectively; a cancelled candidate
may reopen. Let \(a_t\) and \(g_t\) be candidate activity and age during gate
evaluation. The bounded benign-handoff veto is

\vspace{-1.5em}

\begin{equation}
 B_t=
 \ind\!\left[
 a_t=1\land L_t(s_t^-)=1\land g_t\leq G
 \right].
\label{eq:benign_veto}
\end{equation}

\vspace{-0.5em}

Nonlocal support revokes the veto, whereas a locally compatible candidate that
fails to settle remains active beyond \(G\). Entropy and regression
persistence are
\begin{equation}
\scalebox{0.89}{$\displaystyle
\begin{aligned}
 c_t^H
 &=
 \begin{cases}
 c_{t-1}^H+1,
 &t\geq t_{\min}^H
  \land\bar H_t>\tau_H
  \land\neg B_t,\\
 0,&\text{otherwise},
 \end{cases}\\[-0.1em]
 c_t^R
 &=
 \begin{cases}
 c_{t-1}^R+1,
 &\operatorname*{arg\,max}_k p_{t,k}<s_t^-,\\
 0,&\text{otherwise}.
 \end{cases}
\end{aligned}
\label{eq:persistence_counters}
$}
\end{equation}

The deployed gate requests correction when \(\Gamma_t=1\), where
\begin{equation}
 \Gamma_t=
 \ind[c_t^H\geq P_H]
 \ \lor\
 \ind[a_t=1\land g_t>G]
 \ \lor\
 \ind[c_t^R\geq P_R].
\label{eq:gate}
\end{equation}
The branches represent persistent non-benign entropy, unresolved-transition
timeout, and persistent regression. Off-path mass is not a standalone query branch; through
\(L_t\), it governs local transition-candidate and veto
eligibility. Support that violates the local-compatibility
test revokes the veto, while explicit jump and
action-disagreement query branches are disabled. All tests use the pre-update \(s_t^-\), and state updates
occur after evaluating \(\Gamma_t\); confirmation advances the stabilized phase
for \(t+1\), while confirmation, cancellation, or nonlocal routing closes the
candidate.

\subsection{Human Correction, Retraining, and Termination}

When \(\Gamma_t=1\), autonomy stops and the robot holds its measured state.
Correction uses the initial-data record-and-replay pipeline, base-frame
Cartesian-delta convention, schema, and \PolicyRate{} sampling; the full
accepted suffix is retained without fixed trimming. Rejected attempts and
retries are excluded from training and the matched control-time budget but
remain logged. Policies retrain on
\(\mathcal D_i^d\uplus\mathcal C_i^d\), never corrections alone, under frozen
initialization/warm-start, optimizer updates, sampling, normalization,
validation, and checkpoint-selection manifests. The home detector is armed
after the end effector leaves a \HomeArmRadius{} neighborhood of the start and
terminates evaluation at
\(\|\mathbf x_t^{\rm EE}-\mathbf x_{\rm home}^{\rm EE}\|_2\leq\HomeRadius{}\). Controller faults, safety/operator stops,
and acquisition queries remain distinct termination causes.

\begin{figure*}[t]
    \centering
    \includegraphics[width=\textwidth, trim={8pt 175pt 60pt 8pt}, clip]{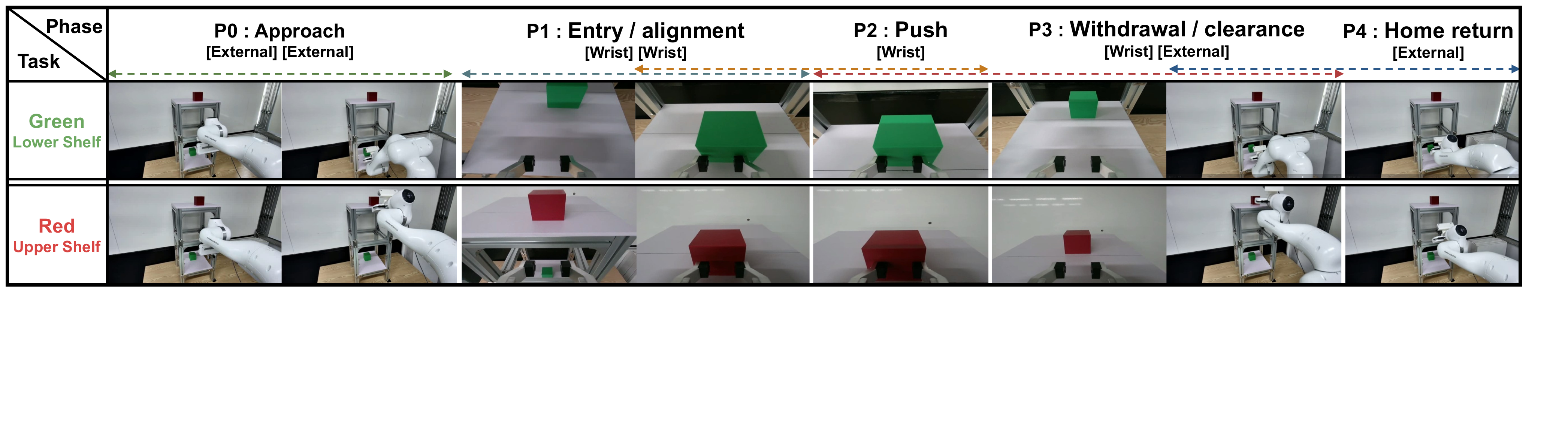}
    \caption{\textbf{Ordered phase structure of the two real-robot pushing tasks.}
\emph{Top:} Green operates on the lower shelf; \emph{bottom:} Red operates on the upper shelf.
Both tasks follow the semantic sequence P0 approach, P1 entry/alignment, P2 push, P3 withdrawal/clearance, and P4 home return.
Representative external and wrist-camera keyframes are shown in execution order.
Adjacent phases share boundary keyframes: the last keyframe of one phase is also used as the onset keyframe of the next phase (P1$\rightarrow$P2, P2$\rightarrow$P3, and P3$\rightarrow$P4).
The panels are illustrative and not time-scaled.}
    \label{fig:phase_structure}
\end{figure*}

\begin{table}[t]
\caption{Active acquisition parameters. G/R denotes Green/Red; all step counts
are at \PolicyRate{}. Deployed native-score thresholds are task-specific,
frozen at 10/20 trajectory-level proposal incidence on disjoint development
sets, and not cross-method comparable. Phase-unaware filtered-\(H\) reuses
\method{} entropy-side settings; exact-\(B\) audit cutpoints are separate and
audit-only.}
\centering
\scriptsize
\setlength{\tabcolsep}{2.7pt}
\resizebox{\columnwidth}{!}{%
\begin{tabular}{@{}ll@{\hspace{5pt}}ll@{}}
\toprule
Quantity & Value & Quantity & Value\\
\midrule
Views / experts &
\NumViews{} / \NumExperts{} &
Chunk / rate &
\ChunkLength{} steps / \PolicyRate{}\\

\(\tau_H\) (G/R), \(W_H\) &
\GreenEntropyThreshold{}/\RedEntropyThreshold{}, \EntropyWindow{} &
\(P_H,t_{\min}^{H}\) &
\EntropyPersistence{}/\MinimumEntropyStep{} steps\\

\(\tau_M,\gamma_{\rm start}\) &
\OffPathThreshold{}, \TransitionStartProb{} &
\(\gamma_{\rm conf}/\gamma_{\rm prev}\) &
\TransitionConfirmNext{}, \TransitionConfirmPrev{}\\

\(P_{\rm conf}/\gamma_{\rm cancel}\) &
\TransitionConfirmSteps{} steps, \TransitionCancelProb{} &
\(P_{\rm cancel}/G\) (G/R) &
\TransitionCancelSteps{} steps,
\GreenTransitionTimeout{}/\RedTransitionTimeout{} steps\\

\(\tau_{\rm ACT}\) (G/R) &
\GreenACTThreshold{}/\RedACTThreshold{} &
\(\tau_{\rm Ens}\) (G/R) &
\GreenEnsembleThreshold{}/\RedEnsembleThreshold{}\\

\(P_R\) &
\RegressionPersistence{} steps &
jump / action query &
disabled / disabled\\
\bottomrule
\end{tabular}}
\label{tab:gate_config}
\end{table}

\section{Experiments}
\label{sec:experiments}

\subsection{Tasks, Data, and Scope}

As illustrated in \Cref{fig:phase_structure}, a Franka Research 3 performs two
phase-structured pushing tasks on a rigid two-level fixture. Green uses the lower shelf; Red follows the same
approach--enter--push--withdraw--return sequence on the upper shelf under
different geometry, visual context, and contact conditions. In both tasks,
\(P0\)--\(P4\) denote approach, entry/alignment, push,
withdrawal/clearance, and home return.
Deployed operating points are calibrated on a separate development-only set
of 20 full-length initial-policy trajectories per task. The 50 Green/50 Red
intervention-free initial-policy archive is disjoint from this calibration
set and is used for the descriptive audit in \Cref{sec:common_trace}; it is
also disjoint from downstream policy evaluation. The initial Green/Red datasets contain 30 demonstrations each (9,956/9,410 frames; 968.3 s total at 20 Hz). The retained MoE-DAgger correction sets contain 2,251/2,551 frames, corresponding to 112.55/127.55 s of active human control.

\subsection{Matched-Budget Methods and Physical Evaluation}
\label{sec:baselines}

Budgets are anchored to the realized \method{} Green/Red active-control times.
Whole suffixes are retained, so the last may cross the target; the stopping rule
and \(\pm\BudgetTolerance{}\) tolerance are fixed before collection. Every
interactive method starts from the same task-specific initial BC checkpoint and
independently acquires learner-state data.

\paragraph{Compared conditions}
\textbf{BC-only} adds no data; \textbf{ExtraDemo-BC} adds full-start
demonstrations matched in active-control time. \textbf{Human-trigger DAgger}
uses score-blind operator takeover when continued autonomy threatens
completion, safety, or interaction quality. \textbf{Phase-matched Random}
samples a frozen phase/progress bin before each rollout and triggers from the
online stabilized router phase without future labels.
\textbf{ACT-Temporal-Disagreement} scores overlapping chunks predicting the
same next motion; \textbf{Ensemble-Disagreement} uses \(N_E=3\)
independently trained same-family policies, one acting and two scoring.
\textbf{Phase-unaware filtered-\(H\)} is the direct same-signal ablation: it
reuses \method{} entropy thresholding, causal averaging, persistence, and
minimum-query settings but removes the transition tracker, adjacent veto,
timeout, and regression without separate incidence recalibration.
\textbf{\method{} (ours)} uses Eq.~\eqref{eq:gate}.

For action-space gates, \(\widetilde{\mathbf a}_{t}^{(j)}\) is predictor
\(j\)'s predicted next-step six-dimensional motion, normalized by frozen initial-BC
statistics; the gripper is excluded. With \(\mathcal P_t\) the overlapping ACT
predictions and \(N_E=3\),
\begin{equation}
\scalebox{0.80}{$\displaystyle
D_t^{\rm ACT}
=\frac{1}{|\mathcal P_t|}\sum_{j\in\mathcal P_t}
\|\widetilde{\mathbf a}_t^{(j)}-\bar{\mathbf a}_t\|_2^2,
D_t^{\rm ens}
=\frac{2\sum_{e<e'}
\|\widetilde{\mathbf a}_t^{(e)}-\widetilde{\mathbf a}_t^{(e')}\|_2^2}
{N_E(N_E-1)}.
$}
\label{eq:disagreement_scores}
\end{equation}

\paragraph{Development-only threshold calibration}
For each task, we use a separate development-only set of 20 full-length
initial-policy trajectories, disjoint from the later 50-trace common-trace
archive. Holding each rule's temporal processing and eligibility logic fixed,
we sweep the native-score threshold for \method{},
ACT-Temporal-Disagreement, and Ensemble-Disagreement and select the value
whose causal replay produces at least one proposal on 10 of the 20
trajectories, or the closest attainable count. No success/failure balancing
or outcome-based subsampling is performed. This calibration equalizes
trajectory-level proposal incidence; it does not optimize failure precision,
critical-window alignment, or retrained-policy performance. Phase-unaware
filtered-\(H\) is not separately recalibrated: it reuses the \method{}
entropy-side settings while removing the transition-aware components. All
thresholds are frozen before added-data acquisition and are not retuned using
physical-evaluation results.

\paragraph{Physical evaluation and reproducibility}
\label{sec:fixed_sets}
For each method and task, two independently seeded checkpoints are evaluated in
25 rollouts each, yielding \(n=50\) per method--task cell and
\TotalRollouts{} rollouts overall. The \(7\,\mathrm{cm}\)-side cube is sampled
\(0\)--\(5\,\mathrm{cm}\) short of the stop line, within
\(\pm4\,\mathrm{cm}\) laterally and \(\pm15^\circ\) in yaw; rollouts are
randomized and interleaved, and poses are logged. All rollouts share the
runtime, controller, sensing/synchronization, temporal ensembling, safety,
home-detector, and episode-cap settings; all added-data policies share
architecture, retraining, sampling, normalization, validation, and checkpoint
selection. Corrections use one logged
takeover--hold--control--acceptance--reset protocol. Active human-control time
excludes offline annotation and uninstrumented attendance, and only
predeclared instrumentation failures are replaced. Each method--task condition
uses one acquired dataset, so the two seeds characterize variation within,
not across, acquisition sets.

\subsection{Common-Trace Acquisition Audit}
\label{sec:common_trace}

Because intervention changes future states, ACT, Ensemble-Disagreement,
phase-unaware entropy, and \method{} are causally replayed on the same
50 Green and 50 Red intervention-free initial-policy traces. For each rule
\(m\) and task \(d\), sweeping the native-score cutpoint assigns one frozen,
outcome-blind rank \(r_{m,d}(i)\) to every archived rollout. Thus,
\(r_{m,d}\) is a method-specific permutation of the same archive rather than
a separate trace set. At selected-rollout budget
\(B\in\{1,\ldots,50\}\), \(c_{m,d}^{(B)}\) retains exactly the first \(B\)
rollouts in that ordering, and \(q_{m,i}(c)\) is the earliest proposal time under
the rule's complete score processing, persistence, and eligibility logic,
not the raw-score peak time. The same ordering is used in all panels of
\Cref{fig:budget_sweep}; \Cref{tab:gate_results} reports \(B=15\) and \(25\), where \(B=15\)
matches the 15 accepted \method{} correction suffixes collected per task. Outcomes and critical
windows are joined only after rankings and first-proposal times are frozen.
Unlike deployed thresholds calibrated on the disjoint 20-trajectory set,
audit-only cutpoints use the held-out 50-trace archive solely for exact-\(B\)
count matching; they are descriptive and not cross-method comparable. The
audit horizon ends before the first offline \(P3\!\rightarrow P4\) onset.

The archive contains 29/36 mutually exclusive Green/Red failures: collision
(3/4), no contact (11/12), and partial push (15/20). Collision is credited only
when the first proposal occurs no later than onset and within the onset or
preceding phase; no-contact and partial-push windows are \(P1\)--\(P2\) and
\(P2\)--\(P3\), respectively. We report aligned failed/proposed yield,
successful-rollout proposals, and per-mode coverage from the first retained
proposal; retrospective timing credit does not establish recoverability.
In \Cref{fig:budget_sweep}, panels~(a,b) count all aligned failures and
panels~(e,f) restrict the same count-matched sweep to collision and no contact.

\subsection{Adjacent-Veto Mechanism Tests}

\paragraph{Shadow component replay}
\label{sec:shadow_ablation}
The common traces are replayed with raw entropy, averaging plus persistence,
the complete rule without the adjacent veto, and the complete rule under the
fixed shadow configuration. All rising edges enter edge and handoff statistics,
whereas alignment uses only the first retained edge; proposal counts are
unmatched and timeout/regression remain bundled. The complete-rule mask is
reused in \Cref{tab:gate_ablation_eligibility} and panels~(c,d) of
\Cref{fig:budget_sweep}. Because no correction or retraining occurs, this
replay diagnoses allocation rather than downstream utility.

\paragraph{Active no-veto test}
\label{sec:active_veto_protocol}
This prospective ablation removes only Eq.~\eqref{eq:benign_veto}; averaging,
persistence, timeout, and regression remain active. Full and no-veto gates
start from the same initial BC checkpoint, independently collect under matched
control time, and use identical retraining and frozen evaluation. Unlike
phase-unaware filtered-\(H\), it preserves the transition model; unlike shadow
replay, it includes collection and retraining. The full row reuses
\Cref{tab:task_results}, and no-veto adds 100 rollouts.

\vspace{-0.5em}

\section{Metrics and Statistical Analysis}
\label{sec:metrics}

\subsection{Downstream and Budget Metrics}

We distinguish collision-free execution of the prescribed physical sequence
from geometric task completion:
\begin{equation}
\begin{aligned}
E_{\rm seq}
&=I_{\rm safe}I_{\rm complete}I_{\rm time},\\[-0.15em]
S_{\rm task}
&=E_{\rm seq}I_{\rm line},\qquad
I_{\rm line}=\ind[m_{\rm push}\leq0].
\end{aligned}
\label{eq:sequence_execution}
\end{equation}
Here, \(I_{\rm safe}=1\) when the rollout completes without collision or a
collision-related protective stop, \(I_{\rm complete}=1\) when the prescribed
physical sequence \(P0\!\rightarrow P1\!\rightarrow P2\!\rightarrow
P3\!\rightarrow P4\) is completed, including withdrawal and return home, and
\(I_{\rm time}=1\) when completion occurs before the episode cap. These
indicators are determined from the synchronized physical rollout outcome, not
from router-head transitions. Thus \(E_{\rm seq}\) does not require stop-line
crossing, whereas \(S_{\rm task}\) additionally requires the cube to cross the
line.
With unit push direction \(\mathbf d_{\rm push}\),
$m_{\rm push}=\mathbf d_{\rm push}^{\top}
(\mathbf p_{\rm goal}-\mathbf p_{\rm cube,rear}(T)).$
Negative values indicate stop-line crossing and positive values the remaining
shortfall; only the sign enters \(S_{\rm task}\), while the magnitude is
descriptive. Active human-control time is the matched supervision budget, and
all non-instrumentation failures remain policy outcomes.

\vspace{-0.25em}

\subsection{Failure-Mode Alignment}

\begin{figure}[t]
    \centering
    \includegraphics[width=0.99\columnwidth]
    {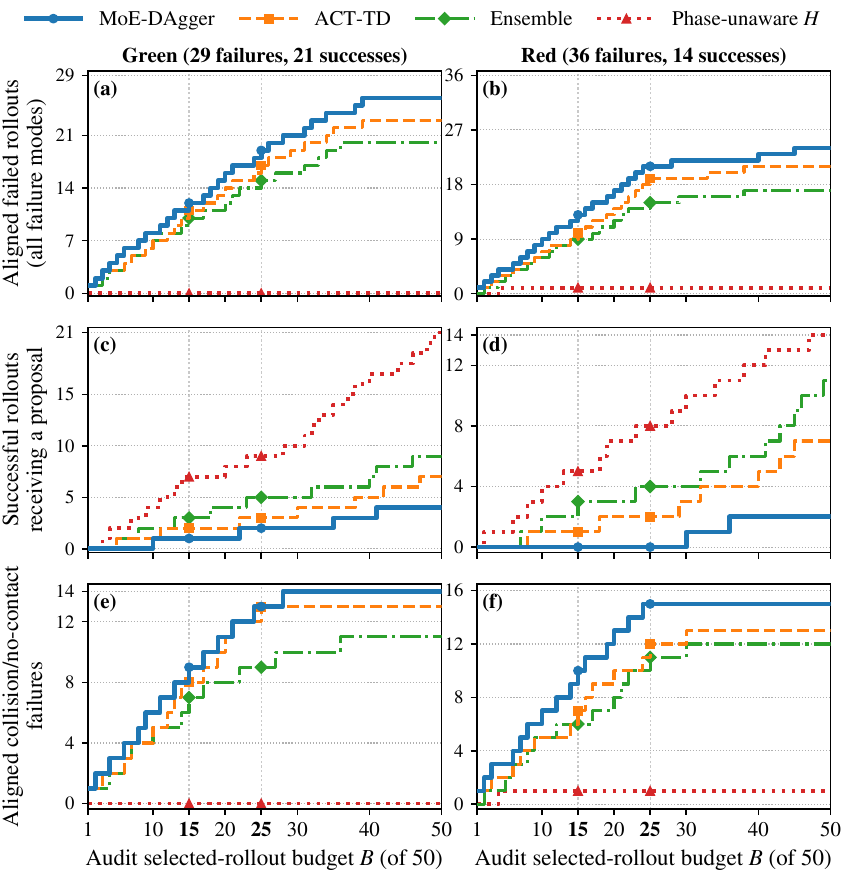}
    \caption{\textbf{Shared-ranking count-matched failure coverage and
fixed-shadow successful-rollout burden.}
Each method--task pair uses one frozen, outcome-blind ordering in all panels.
Panels~(a,b) count aligned failures, and
panels~(e,f) count aligned collision/no-contact failures. Panels~(c,d) apply
the fixed-shadow mask to the same prefixes and count successful rollouts
receiving a proposal, so \(B\) is a rank-prefix size rather than the number
of shadow proposals. At \(B=50\), they therefore count only successful traces
that actually receive a fixed-shadow proposal. Vertical guides mark the
\(B=15,25\) slices in \Cref{tab:gate_results}; higher is better in the
top/bottom rows and lower in the middle row.}
    \label{fig:budget_sweep}
\end{figure}

Using the frozen ordering and audit cutpoints in \Cref{sec:common_trace}, let
\(q_{m,i}^{(B)}=q_{m,i}(c_{m,d}^{(B)})\) be method \(m\)'s earliest retained
proposal on rollout \(i\), with \(q_{m,i}^{(B)}=\varnothing\) when none is
retained. Let \(y_i=1\) denote eventual failure, \(f_i\) its frozen mode, and
\(\mathcal W_{d,f_i}(i)\) the corresponding critical window. With
\(N_{d,f}\) archived failures of mode \(f\), we report
\begin{equation}
    \scalebox{1.0}{$\displaystyle
    \begin{aligned}
    a_{m,i}^{(B)}
    &=\ind\!\left[y_i=1\land
    q_{m,i}^{(B)}\in\mathcal W_{d,f_i}(i)\right],\\
    Y_{m,d}^{(B)}
    &=\frac{1}{B}\sum_i a_{m,i}^{(B)},\\
    N_{m,d}^{\rm succ}(B)
    &=\sum_i\ind\!\left[y_i=0\land q_{m,i}^{(B)}\neq\varnothing\right],\\
    C_{m,d,f}^{(B)}
    &=\frac{1}{N_{d,f}}\sum_{i:f_i=f}
    \ind\!\left[q_{m,i}^{(B)}\in\mathcal W_{d,f}(i)\right].
    \end{aligned}
    $}
    \label{eq:allocation_metrics}
\end{equation}
\(Y_{m,d}^{(B)}\) is the critical-window-aligned failed/proposed yield,
\(N_{m,d}^{\rm succ}(B)\) counts successful traces among the exactly
\(B\) proposal-bearing rollouts, and \(C_{m,d,f}^{(B)}\) is per-mode
coverage.
Consequently, the aligned and successful counts need not sum to \(B\).
All alignment metrics use only the first retained proposal; later proposals
do not repair a mistimed first edge. Higher \(Y\) and \(C\), and lower
\(N^{\rm succ}\), indicate better count-matched allocation on this fixed
archive but do not measure intervention recovery.
Panels~(c,d) of \Cref{fig:budget_sweep} apply a fixed-shadow proposal
mask to the same ordering \(r_{m,d}(i)\). Let
\(\widetilde q_{m,i}\) be the first retained proposal under the fixed
shadow configuration, with
\(\widetilde q_{m,i}=\varnothing\) when no proposal is produced. We plot
\begin{equation}
\scalebox{0.97}{$\displaystyle
\widetilde N_{m,d}^{\rm succ}(B)
=
\sum_i
\ind\!\left[
y_i=0 \land r_{m,d}(i)\le B
\land \widetilde q_{m,i}\neq\varnothing
\right].
$}
\end{equation}

\begin{table*}[t]
\caption{\textbf{Frozen downstream performance under runtime-identical,
matched-supervision evaluation.}
Each method uses two checkpoints from one fixed acquisition set and 25
rollouts per checkpoint and task ($n=50$). Cells show point estimates and 95\%
CIs: Wilson for binary outcomes and pose-stratified rollout bootstrap for
signed margin. Sequence execution does not require stop-line crossing; task success does. Margin is the all-rollout mean ($m_{\rm push}<0$ indicates crossing; more
negative is deeper). Bold/underlining mark best/second-best binary estimates;
ties are retained.}
\centering
\scriptsize
\setlength{\tabcolsep}{0.75pt}
\renewcommand{\arraystretch}{1.02}

\begin{tabular*}{\textwidth}{
@{\extracolsep{\fill}}
l
cc
cc
cc
cc
cc
@{}}
\toprule

&
\multicolumn{2}{c}{Human-control (s)}
&
\multicolumn{2}{c}{Seq. execution (\%)}
&
\multicolumn{2}{c}{Task success (\%)}
&
\multicolumn{2}{c}{Signed margin (cm)}
&
\multicolumn{2}{c}{Collision (\%)}
\\

\cmidrule(lr){2-3}
\cmidrule(lr){4-5}
\cmidrule(lr){6-7}
\cmidrule(lr){8-9}
\cmidrule(lr){10-11}

Method
& G & R
& G & R
& G & R
& G & R
& G & R
\\
\midrule

BC-only
&
0
&
0
&
\pcell{88.0}{76.2}{94.4}
&
\pcell{82.0}{69.2}{90.2}
&
\pcell{52.0}{38.5}{65.2}
&
\pcell{48.0}{34.8}{61.5}
&
\mcell{1.22}{0.981}{1.468}
&
\mcell{1.29}{1.022}{1.612}
&
\pcell{12.0}{5.6}{23.8}
&
\pcell{18.0}{9.8}{30.8}
\\

ExtraDemo-BC
&
114.85
&
128.50
&
\pcell{90.0}{78.6}{95.7}
&
\pcell{82.0}{69.2}{90.2}
&
\pcell{58.0}{44.2}{70.6}
&
\pcell{46.0}{33.0}{59.6}
&
\mcell{0.97}{0.664}{1.397}
&
\mcell{1.27}{1.017}{1.599}
&
\pcell{10.0}{4.3}{21.4}
&
\pcell{18.0}{9.8}{30.8}
\\

Human-trigger DAgger
&
114.45
&
127.95
&
\secondpcell{98.0}{89.5}{99.6}
&
\pcell{92.0}{81.2}{96.8}
&
\pcell{72.0}{58.3}{82.5}
&
\pcell{68.0}{54.2}{79.2}
&
\mcell{-0.18}{-0.398}{0.122}
&
\mcell{0.21}{-0.009}{0.397}
&
\bestpcell{0.0}{0.0}{7.1}
&
\pcell{4.0}{1.1}{13.5}
\\

Phase-matched Random
&
116.20
&
126.45
&
\pcell{92.0}{81.2}{96.8}
&
\pcell{90.0}{78.6}{95.7}
&
\pcell{62.0}{48.2}{74.1}
&
\pcell{60.0}{46.2}{72.4}
&
\mcell{-0.02}{-0.132}{0.105}
&
\mcell{0.63}{0.317}{0.821}
&
\secondpcell{8.0}{3.2}{18.8}
&
\pcell{10.0}{4.3}{21.4}
\\

Phase-unaware filtered-$H$
&
114.85
&
128.35
&
\pcell{86.0}{73.8}{93.0}
&
\pcell{84.0}{71.5}{91.7}
&
\pcell{62.0}{48.2}{74.1}
&
\pcell{52.0}{38.5}{65.2}
&
\mcell{1.05}{0.814}{1.221}
&
\mcell{1.54}{1.129}{1.910}
&
\pcell{14.0}{7.0}{26.2}
&
\pcell{16.0}{8.3}{28.5}
\\

ACT-Temporal-Disagreement
&
115.25
&
128.45
&
\bestpcell{100.0}{92.9}{100.0}
&
\secondpcell{96.0}{86.5}{98.9}
&
\secondpcell{82.0}{69.2}{90.2}
&
\secondpcell{72.0}{58.3}{82.5}
&
\mcell{-1.67}{-2.173}{-0.991}
&
\mcell{-0.81}{-1.215}{-0.542}
&
\bestpcell{0.0}{0.0}{7.1}
&
\bestpcell{0.0}{0.0}{7.1}
\\

Ensemble-Disagreement
&
115.85
&
127.65
&
\secondpcell{98.0}{89.5}{99.6}
&
\secondpcell{96.0}{86.5}{98.9}
&
\bestpcell{84.0}{71.5}{91.7}
&
\pcell{68.0}{54.2}{79.2}
&
\mcell{-1.80}{-2.190}{-1.586}
&
\mcell{-0.41}{-0.928}{-0.007}
&
\bestpcell{0.0}{0.0}{7.1}
&
\secondpcell{2.0}{0.4}{10.5}
\\

\textbf{\method{}}
&
112.55
&
127.55
&
\bestpcell{100.0}{92.9}{100.0}
&
\bestpcell{98.0}{89.5}{99.6}
&
\bestpcell{84.0}{71.5}{91.7}
&
\bestpcell{78.0}{64.8}{87.2}
&
\mcell{-1.98}{-2.280}{-1.521}
&
\mcell{-1.13}{-1.417}{-0.985}
&
\bestpcell{0.0}{0.0}{7.1}
&
\bestpcell{0.0}{0.0}{7.1}
\\

\bottomrule
\end{tabular*}
\label{tab:task_results}
\end{table*}

\begin{table}[t]
\caption{\textbf{Common-trace failure-aligned allocation} on the same 50 Green/50 Red
traces; outcome-blind audit cutpoints retain exactly \(B=15\) or \(25\)
rollouts per method and task (\Cref{sec:common_trace}). ``Align.'': first proposals
inside mode-specific critical windows on failed rollouts; ``Succ.'':
first proposals on successful traces (21 Green/14 Red); other columns:
hits/all archived failures per mode. Bold/underline: best/second-best;
lower is better only for ``Succ.''}

\centering
\footnotesize
\setlength{\tabcolsep}{1.0pt}
\renewcommand{\arraystretch}{1.06}

\newcommand{\alloccell}[2]{%
  \mbox{#1\,{\scriptsize(#2\%)}}}
\newcommand{\allocbest}[2]{%
  \mbox{\textbf{#1}\,{\scriptsize\textbf{(#2\%)}}}}
\newcommand{\allocsecond}[2]{%
  \mbox{\underline{#1}\,{\scriptsize(#2\%)}}}

\begin{tabular*}{\columnwidth}{
@{\extracolsep{\fill}}
lccccc
@{}}
\toprule
&
\multicolumn{2}{c}{Allocation} &
\multicolumn{3}{c}{Mode coverage}\\
\cmidrule(lr){2-3}
\cmidrule(lr){4-6}

Rule &
Align. &
Succ. &
Coll. &
No-cont. &
Partial\\
\midrule

\multicolumn{6}{@{}l}{\textbf{Green, B=15,} (30\% selected fraction)}\\

ACT temporal disagr. &
\allocsecond{11/15}{73.3} &
\allocsecond{2/21}{9.5} &
\textbf{2/3} &
\underline{6/11} &
\textbf{3/15}\\

Ensemble disagr. &
\alloccell{10/15}{66.7} &
\alloccell{3/21}{14.3} &
\textbf{2/3} &
5/11 &
\textbf{3/15}\\

Phase-unaware $H$ &
\alloccell{0/15}{0.0} &
\alloccell{7/21}{33.3} &
\underline{0/3} &
0/11 &
\underline{0/15}\\

\method{} &
\allocbest{12/15}{80.0} &
\allocbest{1/21}{4.8} &
\textbf{2/3} &
\textbf{7/11} &
\textbf{3/15}\\

\midrule

\multicolumn{6}{@{}l}{\textbf{Red, B=15,} (30\% selected fraction)}\\

ACT temporal disagr. &
\allocsecond{10/15}{66.7} &
\allocsecond{1/14}{7.1} &
\underline{2/4} &
\underline{5/12} &
\textbf{3/20}\\

Ensemble disagr. &
\alloccell{9/15}{60.0} &
\alloccell{3/14}{21.4} &
\underline{2/4} &
4/12 &
\textbf{3/20}\\

Phase-unaware $H$ &
\alloccell{1/15}{6.7} &
\alloccell{5/14}{35.7} &
1/4 &
0/12 &
\underline{0/20}\\

\method{} &
\allocbest{13/15}{86.7} &
\allocbest{0/14}{0.0} &
\textbf{3/4} &
\textbf{7/12} &
\textbf{3/20}\\

\midrule

\multicolumn{6}{@{}l}{\textbf{Green, B=25,} (50\% selected fraction)}\\

ACT temporal disagr. &
\allocsecond{17/25}{68.0} &
\allocsecond{3/21}{14.3} &
\textbf{3/3} &
\textbf{10/11} &
\underline{4/15}\\

Ensemble disagr. &
\alloccell{15/25}{60.0} &
\alloccell{5/21}{23.8} &
\textbf{3/3} &
\underline{6/11} &
\textbf{6/15}\\

Phase-unaware $H$ &
\alloccell{0/25}{0.0} &
\alloccell{9/21}{42.9} &
\underline{0/3} &
0/11 &
0/15\\

\method{} &
\allocbest{19/25}{76.0} &
\allocbest{2/21}{9.5} &
\textbf{3/3} &
\textbf{10/11} &
\textbf{6/15}\\

\midrule

\multicolumn{6}{@{}l}{\textbf{Red, B=25,} (50\% selected fraction)}\\

ACT temporal disagr. &
\allocsecond{19/25}{76.0} &
\allocsecond{2/14}{14.3} &
\underline{3/4} &
\underline{9/12} &
\textbf{7/20}\\

Ensemble disagr. &
\alloccell{15/25}{60.0} &
\alloccell{4/14}{28.6} &
\underline{3/4} &
8/12 &
4/20\\

Phase-unaware $H$ &
\alloccell{1/25}{4.0} &
\alloccell{8/14}{57.1} &
1/4 &
0/12 &
0/20\\

\method{} &
\allocbest{21/25}{84.0} &
\allocbest{0/14}{0.0} &
\textbf{4/4} &
\textbf{11/12} &
\underline{6/20}\\

\bottomrule
\end{tabular*}
\label{tab:gate_results}
\end{table}

Unlike \(N_{m,d}^{\rm succ}(B)\), this quantity is not count-matched:
\(B\) denotes the common rank-prefix size, while the number of
fixed-shadow proposals within that prefix may be smaller than \(B\).
At \(B=50\), it equals the total number of successful traces that actually
receive a proposal under the fixed shadow rule.
The complete-rule row of \Cref{tab:gate_ablation_eligibility} uses the
same \(\widetilde q_{m,i}\). There, ``Prop.'' is the number of rollouts
with any retained rising edge, and ``fail/prop.'' credits a failed
rollout only when its first retained edge lies in
\(\mathcal W_{d,f_i}(i)\). Handoff overlap uses all retained edges and is
a transition-confounding diagnostic rather than a calibrated
false-positive rate.

\subsection{Evidence and Statistical Analysis}

Rollouts are the observational units; task success is the primary
endpoint, with sequence execution and collision complementary and
signed margin descriptive. The mechanism-specific comparison is
\method{} versus phase-unaware filtered-\(H\). Per method--task cell
(\(n=50\)), binary rates use Wilson 95\% CIs, pairwise differences
unadjusted Newcombe 95\% CIs, and margin a rollout-level bootstrap;
common-trace values are exact archive counts without CIs.

\begin{figure*}[h]
    \centering
    \includegraphics[width=\textwidth, trim={14pt 705pt 67pt 27pt}, clip]{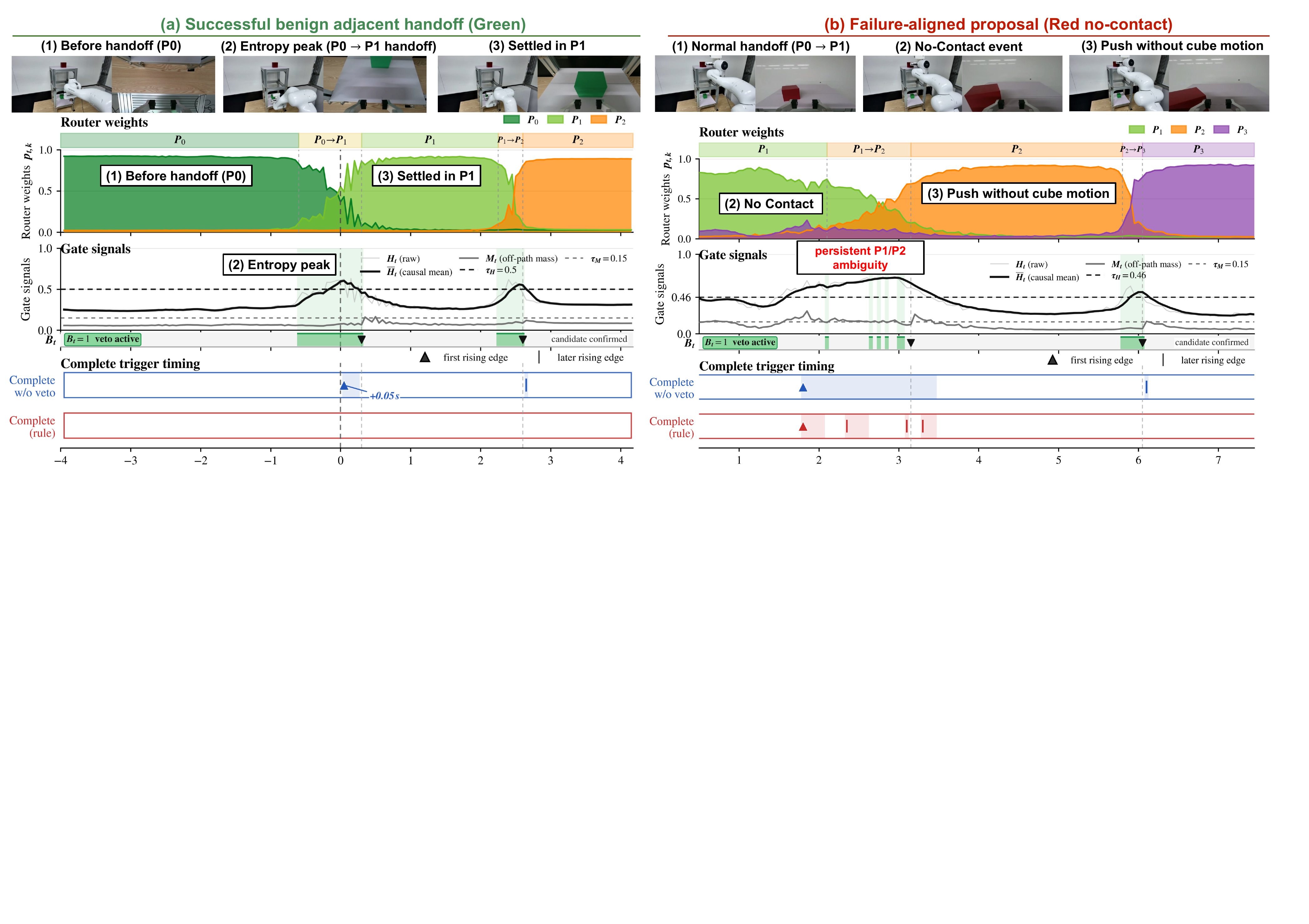}
    \caption{\textbf{Illustrative adjacent-handoff veto behavior under fixed-shadow replay.}
(a) On a successful Green trace, no-veto produces its first proposal edge
\(0.05\,\mathrm{s}\) after the offline \(P0\!\rightarrow P1\) handoff, whereas the
full rule suppresses queries during the bounded handoff and produces no retained edge in the
shown interval. (b) On a Red no-contact failure, the full rule retains a
failure-aligned proposal edge during persistent \(P1/P2\) ambiguity. Curves show
router weights \(p_{t,k}\), raw and causal-mean entropy \(H_t,\bar{H}_t\), off-path
mass \(M_t\), and veto state \(B_t\); the lower rows show no-veto and full-rule
shadow edges. Triangles mark the audit-scored first edge and ticks later edges.
Phase bands and frame annotations are offline visualizations, not runtime inputs.
Each trace uses its task-specific deployed threshold and the fixed-shadow
configuration of \Cref{tab:gate_ablation_eligibility}; no audit-only exact-\(B\)
cutpoint is applied.}
    \label{fig:veto_trace}
\end{figure*}

\section{Results}
\label{sec:results}

\subsection{Matched-Budget Downstream Utility}
\label{sec:frozen_downstream}

Under nearly identical active human-control time, \method{} achieves
84.0/78.0\% task success, 100.0/98.0\% sequence execution, and no observed
collisions on Green/Red (\Cref{tab:task_results}). Relative to the direct
same-signal phase-unaware gate, this is \(+22/+26\) percentage points in task
success (95\% CI \([4.6,37.8]/[7.3,42.4]\)), \(+14/+14\) in sequence execution
(\([4.0,26.2]/[2.6,26.6]\)), and \(-14/-16\) in collision
(\([-26.2,-4.0]/[-28.5,-5.5]\)).
ExtraDemo-BC, Human-trigger DAgger, and Phase-matched Random---controlling for
full-trajectory supervision, score-blind human selection, and phase-matched
online data---reach 58.0/46.0\%, 72.0/68.0\%, and 62.0/60.0\% task success,
respectively. These results indicate that those factors alone do not account
for the gain under matched active-control time. ACT-TD remains competitive at
82.0/72.0\% task success with no observed collisions. The task-success
differences (\method{} minus ACT-TD) are \(+2/+6\) percentage points, with
95\% CIs spanning zero (\([-13.0,16.9]/[-10.9,22.5]\)).

\begin{table}[t]
\caption{\textbf{Shadow gate-component audit.}
All variants are replayed on the same 50 Green and 50 Red
intervention-free initial-policy traces used for the common-trace audit,
using the task-specific deployed entropy threshold frozen on a separate,
disjoint 20-trajectory development set for each task. Unlike
\Cref{fig:budget_sweep} (panels~(a,b,e,f)), no audit-only exact-\(B\) cutpoint is imposed;
all rising edges within the frozen audit horizon are retained.
``Prop.'' counts rollouts with any retained edge, ``aligned fail/prop.''
is the fraction of proposal-bearing rollouts that eventually fail and whose
first retained edge lies in the corresponding mode-specific window defined in
\Cref{sec:common_trace}, and ``handoff ovlp.'' is the fraction of
adjacent handoffs intersected by any retained edge.}
\centering
\scriptsize
\setlength{\tabcolsep}{2.2pt}
\renewcommand{\arraystretch}{1.08}

\resizebox{\columnwidth}{!}{%
\begin{tabular}{@{}lcccc@{}}
\toprule
Variant &
Edges &
Prop. &
\shortstack{Aligned\\fail/prop.} &
Handoff ovlp.\\
\midrule

\multicolumn{5}{@{}l}{%
\textit{A. Green}}\\

Raw \(H_t\) &
314 &
50 / 50 (100.0\%) &
0 / 50 (0.0\%) &
147 / 147 (100.0\%)\\

Avg. \(+\) persist. &
170 &
50 / 50 (100.0\%) &
0 / 50 (0.0\%) &
97 / 147 (66.0\%)\\

Complete w/o veto &
170 &
50 / 50 (100.0\%) &
0 / 50 (0.0\%) &
97 / 147 (66.0\%)\\

Complete rule (shadow) &
\textbf{36} &
\textbf{32 / 50 (64.0\%)} &
\textbf{24 / 32 (75.0\%)} &
\textbf{0 / 147 (0.0\%)}\\

\midrule

\multicolumn{5}{@{}l}{%
\textit{B. Red}}\\

Raw \(H_t\) &
250 &
50 / 50 (100.0\%) &
0 / 50 (0.0\%) &
136 / 143 (95.1\%)\\

Avg. \(+\) persist. &
106 &
50 / 50 (100.0\%) &
1 / 50 (2.0\%) &
71 / 143 (49.7\%)\\

Complete w/o veto &
104 &
50 / 50 (100.0\%) &
1 / 50 (2.0\%) &
71 / 143 (49.7\%)\\

Complete rule (shadow) &
\textbf{34} &
\textbf{27 / 50 (54.0\%)} &
\textbf{22 / 27 (81.5\%)} &
\textbf{3 / 143 (2.1\%)}\\

\bottomrule
\end{tabular}}
\label{tab:gate_ablation_eligibility}
\end{table}

\subsection{Failure-Mode-Aligned Common-Trace Allocation}
\label{sec:common_trace_results}

\vspace{-0.25em}

Pooled across the two tasks, \method{} attains critical-window-aligned
failed/proposed yields of \TraceAlignedRateOurs{} at \(B=15\) and \(25\),
respectively, compared with \TraceAlignedRateACT{} for ACT-TD,
\TraceAlignedRateEnsemble{} for the ensemble, and
\TraceAlignedRatePhaseUnaware{} for phase-unaware entropy
(\Cref{tab:gate_results}). It proposes correction on only 1 and 2 of the 35
successful traces at \(B=15\) and \(25\), respectively, indicating that
transition awareness shifts proposals away from benign successful-rollout
handoffs.

Across the full sweep in \Cref{fig:budget_sweep}, at \(B=50\) \method{}
aligns 26/29 Green and 24/36 Red failures, including 14/14 and 15/16
collision/no-contact cases. The remaining misses are concentrated in partial
push, where coverage is 12/15 on Green and 9/20 on Red. Phase-unaware entropy aligns only 1 of 18 and
1 of 33 selected failed rollouts at \(B=15\) and \(25\), respectively. These counts show that selecting failed rollouts alone does not ensure that first proposals fall within the corresponding critical windows.
ACT is stronger on Red partial push at \(B=25\) (7/20 versus 6/20), so the
advantage is not uniform across all failure modes.

\begin{table}[t]
\caption{\textbf{Active adjacent-veto ablation.}
No-veto removes only Eq.~\eqref{eq:benign_veto}; both variants independently
collect under matched control time with identical retraining/evaluation. Veto q./rollout denotes the percentage of acquisition rollouts whose query
onset meets the counterfactual veto condition. Effects are Full minus No-veto; binary effects are percentage
points.}
\centering
\scriptsize
\setlength{\tabcolsep}{2.0pt}
\renewcommand{\arraystretch}{1.08}

\resizebox{\columnwidth}{!}{%
\begin{tabular}{@{}lccccc@{}}
\toprule
Acquisition gate &
\shortstack{Human-control\\(s)} &
\shortstack{Veto q.\\{/rollout (\%)}} &
\shortstack{Seq.\\execution (\%)} &
\shortstack{Task\\success (\%)} &
\shortstack{Collision\\(\%)}\\
\midrule

No adjacent veto &
114.70 / 130.20 &
92.0 / 50.0 &
90.0 / 84.0 &
68.0 / 60.0 &
10.0 / 16.0 \\

Full \method{} &
112.55 / 127.55 &
0.0 / 0.0 &
100.0 / 98.0 &
84.0 / 78.0 &
0.0 / 0.0 \\

\(\Delta\) (Full -- No-veto) &
\(-2.15/-2.65\) &
\(-92.0/-50.0\) &
\(+10.0/+14.0\) &
\(+16.0/+18.0\) &
\(-10.0/-16.0\) \\

\bottomrule
\end{tabular}%
}
\label{tab:active_veto}
\end{table}

\subsection{Adjacent-Veto Mechanism Tests}

\paragraph{Shadow replay}
Averaging and persistence reduce raw proposal edges from 314 to 170 on Green
and from 250 to 106 on Red, yet without the adjacent veto every trace still
receives a proposal, with first edges concentrated near \(P0\) or the nominal
\(P0\!\rightarrow P1\) handoff. The complete veto reduces edge counts to
36/34, proposal incidence to 32/27 of 50, and handoff overlap from
66.0/49.7\% to 0.0/2.1\%, while increasing first-proposal alignment to
24/32 (75.0\%) and 22/27 (81.5\%)
(\Cref{tab:gate_ablation_eligibility}). \Cref{fig:veto_trace} visualizes this
contrast: the full rule suppresses a handoff-confounded proposal edge on a
successful Green trace but retains a failure-aligned proposal edge during
persistent \(P1/P2\) ambiguity on a Red no-contact trace.
The unmatched aligned counts are lower than the \(B=50\) count-matched counts
in \Cref{fig:budget_sweep}, even though both analyses use the same traces. The
count-matched sweep lowers the audit cutpoint until all 50 archived traces are
proposal-bearing, whereas the fixed-shadow complete rule produces proposals on
only 32/27 of those same Green/Red traces. This supports selective suppression
of handoff-confounded proposals rather than temporal smoothing alone.

\paragraph{Active no-veto test}
Relative to no-veto, the full gate reduces the incidence of queries satisfying
the veto condition by 92/50 percentage points and improves sequence execution
by 10/14 percentage points (\([0.9,21.4]/[2.6,26.6]\)). The collision-rate
differences (full minus no-veto) are \(-10/-16\) percentage points
(\([-21.4,-0.9]/[-28.5,-5.5]\)). Task success improves by 16/18 percentage
points, although the intervals for these differences include zero
(\([-0.8,31.8]/[-0.1,34.6]\)); the sequence-execution and collision-rate
intervals exclude zero in both tasks.

\section{Discussion}
\label{sec:discussion}

\paragraph{What the evidence supports}
The active phase-unaware comparison tests downstream utility with fixed
entropy-side settings, whereas the common-trace audit tests count-matched
first-proposal allocation. Together, consistent with Proposition~1, they show
that phase topology changes both rollout selection and proposal timing because
equal entropy can represent a settling adjacent handoff or nonlocal/regressive
support. The contribution is topology-aware allocation rather than universal
dominance over phase-unaware uncertainty.

\paragraph{Interpretation and veto tradeoff}
The five heads are phase-anchored action predictors rather than independently
reusable skills, so the result should be interpreted as an acquisition effect
rather than skill discovery or multitask expert reuse. The adjacent veto
deliberately trades proposal incidence for reduced interruption at compatible
handoffs. Shadow replay isolates this allocation effect, while the active
ablation shows its downstream safety and execution consequences under matched
control time.

\paragraph{Scope and limitations}
The study uses one robot and fixture, separate task-specific policies, and two
geometric instances of one pushing family; multitask reuse, learned or
nonlinear phase graphs, unsupervised skill discovery, other robots, and
open-world settings remain untested. Each method--task condition has one
acquired dataset, so two training seeds and repeated rollouts do not quantify
acquisition-set variability. The matched budget captures active human-control time rather than total
operator workload; it excludes offline phase annotation and uninstrumented
attendance.
Retrospective critical windows do not establish
that a credited intervention would recover a rollout, collision coverage
contains only 3 Green and 4 Red failures, and partial-push coverage remains
limited. One aggregation round establishes neither convergence nor an optimal stopping
rule. Extending to larger task- or skill-level MoEs may require hybrid gates:
the router may concentrate on one expert and remain low-entropy even when the
resulting action predictions exhibit high disagreement, motivating fusion with
complementary signals such as ACT temporal disagreement.

\section{Conclusion}
\label{sec:conclusion}

\method{} conditions persistent soft-MoE router entropy on legal phase support
and temporal settling to determine when to request human correction. Under
matched active human-control time, it improves task success by
\HeadlineTaskSuccess{} over phase-unaware entropy and records no observed
collisions in 50 evaluation rollouts per task. Common-trace allocation and
adjacent-veto ablations show that this gain is associated with suppressing
benign handoff proposals and reallocating correction toward
failure-relevant windows. Coverage is strongest for collision and no-contact,
while partial-push failures remain the main unresolved mode.
These results demonstrate the practical value of known phase topology as an
acquisition prior for selective DAgger in structured real-robot manipulation.


\bibliographystyle{IEEEtran}
\bibliography{references}

\end{document}